\pdfoutput=1 
\RequirePackage{fix-cm}
\documentclass[smallcondensed]{svjour3}     
\smartqed  
\usepackage{graphicx}
\usepackage{multirow}
\usepackage{cite}
\usepackage{algorithm, algorithmic}
\usepackage{amsfonts}
\usepackage{amsopn}
\usepackage{amstext}
\usepackage{amsbsy}
\usepackage{subfigure}
\usepackage{amsmath}
\usepackage{array}
\usepackage{indentfirst}
\usepackage{amssymb, bm}
\usepackage{ragged2e}
\begin{document}

\title{Human Action Recognition Based on Multi-scale Feature Maps from Depth Video Sequences
}

\titlerunning{Human Action Recognition Based on LP-DMI}        

\author{Chang Li \and  Qian Huang* \and Xing Li \and Qianhan Wu 
}


\institute{
	Qian Huang \at
	School of Computer and Information, Hohai University, Hohai University, Nanjing, China.\\
	Tel.: +86-15062245312\\
	\email{huangqian@hhu.edu.cn}         
	\and
	Chang Li\at
	School of Computer and Information, Hohai University, Hohai University, Nanjing, China.\\
	Tel.: +86-15951730907\\
	\email{lichang@hhu.edu.en}           
	\and
	Xing Li\at
	School of Computer and Information, Hohai University, Hohai University, Nanjing, China.\\
	\email{lixing@hhu.edu.cn}           
	\and
	Qianhan Wu\at
	School of Computer and Information, Hohai University, Hohai University, Nanjing, China.\\
	\email{wuqianhan@hhu.edu.cn} 
}
\date{Received: date / Accepted: date}

\maketitle
\begin{abstract}
Human action recognition is an active research area in computer vision. Although great process has been made, previous methods mostly recognize actions based on depth data at only one scale, and thus they often neglect multi-scale features that provide additional information action recognition in practical application scenarios. In this paper, we present a novel framework focusing on multi-scale motion information to recognize human actions from depth video sequences. We propose a multi-scale feature map called Laplacian pyramid depth motion images(LP-DMI). We employ depth motion images (DMI) as the templates to generate the multi-scale static representation of actions. Then, we caculate LP-DMI to enhance multi-scale dynamic information of motions and reduces redundant static information in human bodies. We further extract the multi-granularity descriptor called LP-DMI-HOG to provide more discriminative features. Finally, we utilize extreme learning machine (ELM) for action classification. The proposed method yeilds the recognition accuracy of 93.41\%, 85.12\%, 91.94\% on public MSRAction3D dataset, UTD-MHAD and DHA dataset. Through extensive experiments, we prove that our method outperforms state-of-the-art benchmarks.

\end{abstract}

\section{Introduction}
%
\label{Introduction}
Human action recognition is a hot topic in computer vision, which aims to automatically interpret the semantic information conveyed by human actions and the human interactions with the external environment. It has many real-world applications, such as security monitoring , intelligent human-computer interaction, smart home, and elderly healthcare etc.\cite{RF1,RF2,RF3,RF4,RF5}. However, this task is still challenging because of problems like illumination, occlusion, varying spatio-temporal scale, clothing, and viewing angles. 
\par Initially, action recognition technology was mainly based on RGB data acquired by ordinary cameras \cite{RF7,RF8,RF9}. However, RGB data is tempted with external factors, such as shooting environment, lighting, and wearing texture, which has limited the development of action recognition. With the introduction of low cost depth sensors, such as Microsoft Kinect, ASUS Xtion and other RGB-D sensors, major breakthroughs have been made in human action recognition. Compared to traditional RGB data, depth video sequences provide 3D structure data of actions. The pixels of depth maps describe the distance between the surface of objects and sensors \cite{RF11}. This depth information provides convenience for segmenting the foreground person and eliminates the interference caused by complex backgrounds. Consequently, depth maps have better invariance to illumination and texture changes. Recently, a variety of methods have been investigated to describe depth video data for action recognition \cite{RF12,RF13,RF14,RF15}. However, these descriptors all ignore the multi-scale data of motions. In actual application scenarios e.g. security monitoring, recognizing moving human is a tricky task, which contains abundant spatial information in different scales. Therefore, we motivate to represent motions by a muti-scale model to capture more discriminative features.
\par In this paper, we present a novel human action recognition framework with a multi-scale mechanism illustrated in Fig. 1. We project the depth video sequences onto three orthogonal Cartesian planes to obtain three-view depth motion images (DMI) which constitutes the 3D action model. After that, we apply the Gaussian pyramid to simulate the scale changes of human eyes and obtain the static multi-scale representation of actions. Then, we construct Laplacian pyramids to generate the compact feature map LP-DMI which enhances the dynamic multi-scale information for action recognition. 
\begin{figure*}[h]
	\centering{\includegraphics[width=1.00\textwidth]{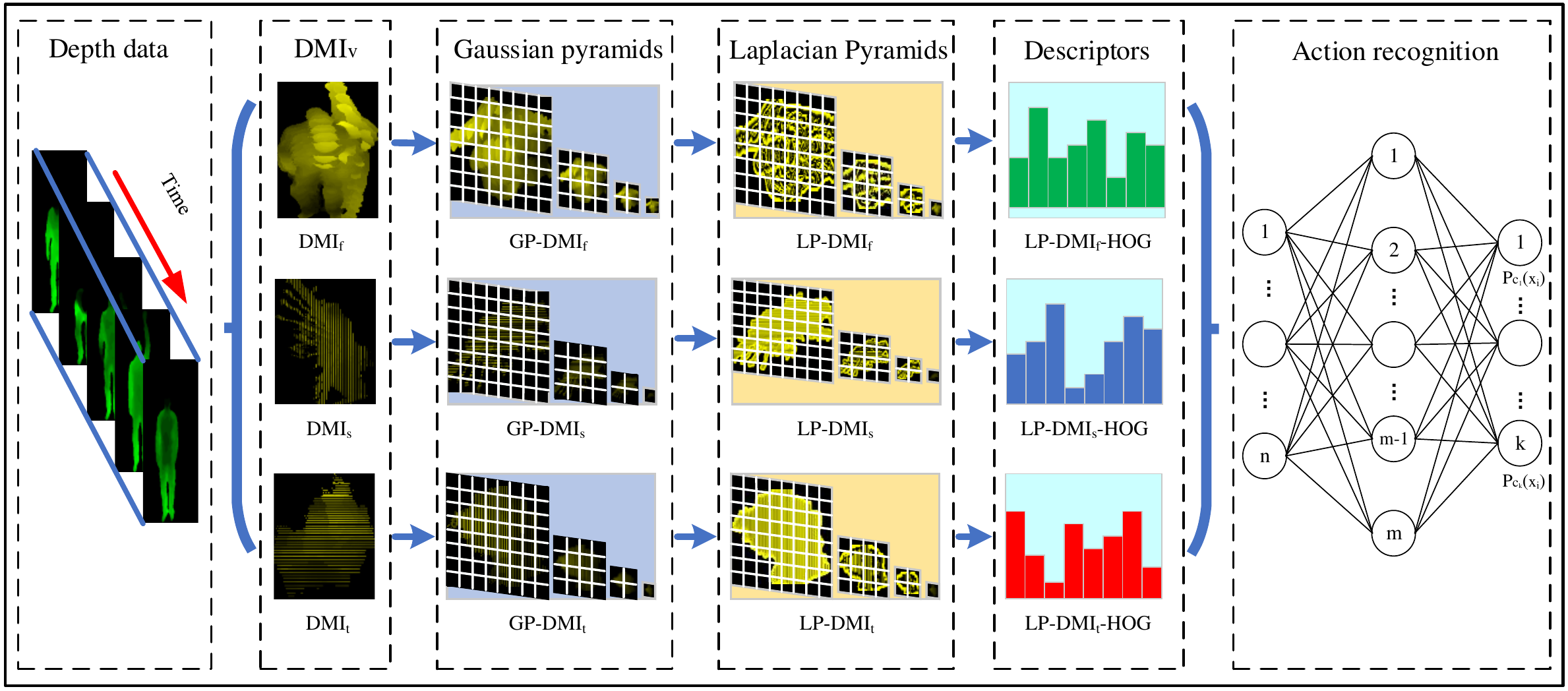}}
	\caption{Framework of our proposed human action recognition method.\label{fig.1}}
\end{figure*}
After that, the LP-DMI-HOG capturing multi-granularity motion features can be extracted following the pyramid structure. Finally, we employ ELM to classify actions. Specifically, the main contributions of this article are summarized as follows:
\begin{enumerate}
\item[1)] We study a compact multi-scale feature map based on depth data called LP-DMI. Due to its superiority of enhancing multi-scale dynamic information of actions, the proposed feature map outperforms other existing maps. Moreover, some redundant static information inside the body is excluded simultaneously.
\item[2)] We introduce a feature extraction scheme according to the hierarchical structure of Laplacian pyramids. We extract HOG features and cascade them as LP-DMI-HOG. This descriptor captures multi-granularity features therefore it is more discriminative than others.
\item[3)] We propose a multi-scale human action recognition framework based on the depth video sequences. We generate compact multi-scale feature maps through the Laplacian pyramid of three-view DMI and then extract multi-granularity features. After that, we use extreme learning machine for action classification.
\item[4)] We conduct experiments on the public MSRAction3D, UTD-MHAD and DHA dataset, and the experimental results demonstrate that our method surpasses state-of-the-art benchmarks.
\end{enumerate} 
\par The rest of this article is organized as follows. Section 2 reviews the previous work related to ours. In Section 3, the proposed method is presented in detail, including building Laplacian pyramids of DMI, extracting LP-DMI-HOG feature and action classification. Section 4 discusses the experimental setttings and results compared to other human action recognition methods. At last, the conclusions of this paper are drawn in Section 5.
\section{Related work}
\label{sec:1}
\par  According to the type of input data, the human action recognition technologies consist of RGB video based methods\cite{RF7,RF8,RF9}, depth video based methods\cite{RF21,RF19,RF62,RF66}, depth skeleton based methods\cite{RF6,RF49,RF40}, and multi-modal data fusion based methods \cite{RF60,RF61,RF63}. Due to the convenience of data acquisition and invariance to illumination and texture changes, many researchers focus on depth video based methods. Generally, depth video based human action recognition aproaches contains three steps: computing depth feature maps from depth video sequences, generating feature descriptors and action clasification. For higher accuracy, tremendous effort has been made to investigate representation and feature extraction strategy for human action recognition.Bobick and Davis \cite{RF13} introduced a view-based approach on the basis of a temporal template that contains two component versions: the presence and recency of motion in sequence. They computed motion energy images (MEI) and motion history images (MHI) to model spatial and temporal characteristics of human actions. Mohammad $et$ $al$. \cite{RF11} utilized the static history images (SHI) as the complementary components of MHI. Motivated by MHI and MEI, Yang $et$ $al$. \cite{RF19} projected each depth frame onto three orthogonal Cartesian planes, then the subtraction operations between successive projections were carried out to obtain depth motion maps (DMM). On the contrary to DMM, Kamel $et$ $al$. \cite{RF12} investigated the depth motion images (DMI) in which the pixel value is the minimum value of the position of the same pixels over time to describe the overall action appearance from the front view. Since the DMM fails to recognize two actions with reverse temporal orders, Elmadany $et$ $al$. \cite{RF21} divided the depth video sequences into multiple partitions with the equal number of frames. Then they constructed the hierarchical pyramid depth motion maps (HP-DMM) so as to capture more detailed information of human movements.
\par Based on the depth feature maps above, many descriptors have been studied for human action recognition. The histogram of oriented gradients (HOG) \cite{RF22}, the local binary pattern (LBP) \cite{RF45}, and other shape and texture features \cite{RF23} were calculated from DMM for more accurate description. Oreifej and Liu \cite{RF25} introduced the histogram of oriented 4D normals (HON4D) in order to describe the action in 4D space, including depth, spatial, and time coordinates. Li $et$ $al$. \cite{RF51} introduced Local Ternary Pattern (LTP) as an image filter for DMMs and applied CNN to classify corresponding LTP-encoded images. Tian $et$ $al$. \cite{RF26} employed Harris detector and local HOG descriptor on MHI for action recognition and detection. Furthermore, Gu $et$ $al$. \cite{RF14} selected ResNet-101 as the deep learning model and feed it with MHI. Aly $et$ $al$. \cite{RF50} calculated global and local features using Zernike moments with different polynomial orders to represent global and local motion patterns respectively.Kamel $et$ $al$. \cite{RF12} presented a feature fusion method for human action recognition from DMI and moving joints descriptor (MJD) data using convolutional neural networks (CNN). Mohammad $et$ $al$. \cite{RF11} extracted the gradient local auto-correlations (GLAC) features from the MHI along with SHI to represent the movements. Chen $et$ $al$. \cite{RF28} computed GLAC features based on DMM and put them into the extreme learning machine for activity recognition. Space time occupancy patterns (STOP) was proposed by Vieira $et$ $al$. \cite{RF44} in which space and temporal axes are divided into several partitions for each sequence. The Bag of Angles (BoA) for skeleton sequence and another descriptor called Hierarchical pyramid DMM deep convolutional neural network (HP-DMM-CNN) for depth videos were presented in \cite{RF21}. 
\par In addition, some new methods have emerged in the latest work. Sun $et$ $al$. \cite{RF47} presented a global and local histogram representation model using the joint displacement between the current frame and the first frame, and the joint displacement between pairwise fixed-skip frames. Then naive-bayes nearest-neighbor (NBNN) and sparse representation-based classifier (SRC) were applied for action recognition. Ahmad $et$ $al$. \cite{RF46} fed feature maps into the CNN architecture rather than using any conventional method. Ulteriorly, Trelinski $et$ $al$. \cite{RF48} concatenated handcrafted features and action-specific CNN-based features together as the action feature vectors and utilize support vector machines (SVM) for classification. Li $et$ $al$. \cite{RF52}  generated 3D body mask and then formed the depth spatial-temporal maps (DSTMs) which provided compact global spatial and temporal information of human motions. In addition, they compared the results of various classifiers such as K-nearest neighbor (KNN), random forest (BF), and gaussian naive bayesian model (GNBM). Wei $et$ $al$. \cite{RF49}  modeled human actions with a hierarchical graph in which the depth video sequence was represented as sequential atomic actions. Every atomic action was denoted as a composite latent state consisted by a latent semantic attribute and a latent geometric attribute. In their work, hidden markov model (HMM) with AdaBoost, dynamic temporal warping, and recurrent neural network (RNN) were employed.However, the methods above fail to capture the multi-scale features for action recognition, and thus have poor robustness. Recently, more attention has been paid to multi-scale motion information. Ji $et$ $al$. \cite{RF43} embedded the skeleton information into depth feature maps to divide the human body into several parts. The surface normals of local motion part sequence were partitioned into different space-time cells to obtain local spatio-temporal scaled pyramid which was applied to extract local feature representation. Yao $et$ $al$. \cite{RF20} studied parallel pair discriminant correlation analysis (PPDCA) to fuse the multi-temporal-scale information with a lower dimension. However, the multitemporal-scale in this method means features related to different numbers of frames. These solutions obtain multi-scale information by different number of frames and cells or various sampling rate, which is only the scale change in the temporal level in essence. In this paper, we present a multi-scale method based on the Scale-space theory in \cite{RF29}. Note that rather than realize multi-temporal scale, we focus on spatial multi-scale of feature maps to tackle the problem of complex model representation and low implementation efficiency.
\section{PROPOSED METHOD FOR HUMAN ACTION RECOGNITION}
A typical action contains characteristic information in different scales, and it can be represented by the structured multi-scale features. Learning the information in single spatial scale is deficient to provide discriminative feature sufficiently for human action recognition. Therefore, we propose a novel method to represent the multi-scale motion information. Our approach contains three sub-modules, including calculating feature map LP-DMI, extracting LP-DMI-HOG descriptor, and classifying actions by ELM.
\begin{figure*}[!t]
	\centering{\includegraphics[width=1.00\textwidth]{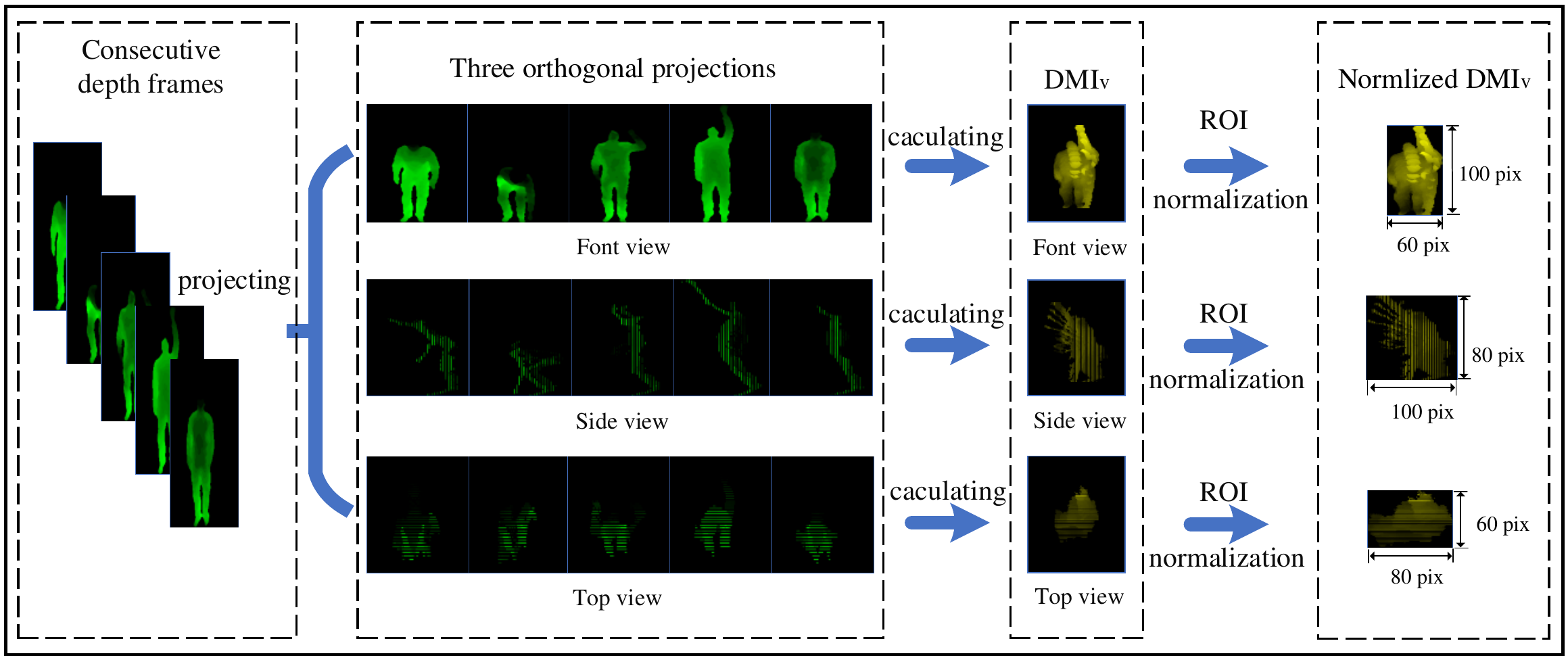}}
	\caption{The process of calculating {$\rm DMI_{v}$} from depth video sequences\label{fig2}}
\end{figure*} 
\subsection{Calculating feature map LP-DMI}
With the advent of depth cameras, a lot of approaches have been introduced based on the depth data for human action recognition. Each frame of the depth camera records a snapshot of the action at a certain point in time. In general, DMI is considered as an effective representation of depth video sequences. It captures not only the overall appearance of actions but also the depth changes in the moving parts. In this paper, we project the depth frames obtained by the depth camera onto three orthogonal Cartesian coordinate planes, thus each 3D depth frame generates three 2D maps. We record them as $\operatorname{map}_{v}(v \in\{f, s, t\})$ corresponding to the front, side, and top view respectively. The pixel value of DMI is the minimum value of the same spatial position of the depth maps. The three-view DMI of a depth video sequences with $N$ frames can be calculated using the following equation.
\begin{equation}\begin{array}{l}
D M I_{v}(i, j) =255-\min \left(\operatorname{map}_{v}(i, j, t)\right) ,\\
\qquad \qquad \forall t  \in[k, \ldots,(k+N-1)]
\end{array}\end{equation}
where $\operatorname{map}_{v}(i, j, t)$ is the pixel value of $(i,j)$ position of 2D map at time $t$ from the perspective of $v$. $k$ represents the index of the frame. The maps are processed by dividing each pixel value by the maximum value of all the pixels contained in the image for normalization. We crop the region of interest (ROI) in DMI to exclude excess black pixels. This normalization contributes to eliminating intra-class differences and reducing the nuisances caused by body shape and motion amplitude. The generative process of DMI is depicted in Fig. 2.
\begin{figure}[!t]
	\centering{\includegraphics[width=0.75\textwidth]{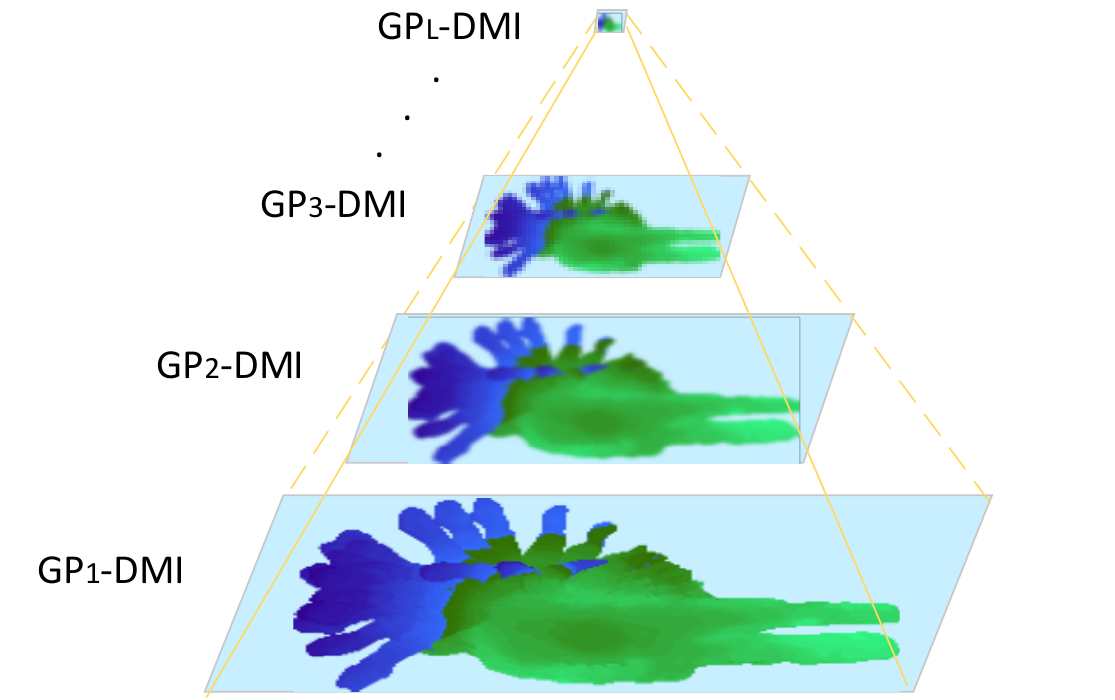}}
	\caption{The hierarchical structure of GP-DMI\label{fig3}}
\end{figure}
\par However, DMI simply reflects spatial information of actions in single scale. In order to capture multi-scale changes of human motions, we adopt the Gaussian pyramid transform which has been demonstrated the practicability in increasing scale diversity \cite{RF30,RF32}. As shown in Fig. 3, we acquire a cluster of multi-scale feature images shaped like several pyramids. We stipulate that the number of layers goes up in a bottom-up manner. $G_l$ is used to represent the image of $l_{th}$ layer of a Gaussian pyramid, that is to say, the size of $G_{l+1}$ is smaller than that of the $G_l$. we need to perform Gaussian kernel convolution and downsampling on the $G_l$ to produce $G_{l+1}$. Mathematically, the gray value corresponding to the $(i, j)$ position of $G_l$  can be formulated as:
\begin{gather}
G_{l}(i, j)=\sum_{m=-c}^{c}\sum_{n=-c}^{c} \varpi(m, n) \otimes G_{l-1}\left(2 i+m, 2 j+n\right), \notag \\
\left(1 \leq l \leq L, 0 \leq i \leq R_{l}, 0 \leq j \leq C_{l}\right)
\end{gather}
where $\otimes$ is a convolution operator and $L$ is the total number of layers in every Gaussian pyramid. $(m, n)$ is the position of the convolution kernel. $R_l$ and $C_l$ are the number of rows and columns relative to the $l_{th}$ layer image of the Gaussian pyramid. $c$ determines the size of $\varpi$ and $\varpi$ is a Gaussian window of size  $(2c+1)\times(2c+1)$ satisfying the following formula:
\begin{equation}\varpi(m, n)=\frac{1}{2 \pi \sigma^{2}} e^{-\left(m^{2}+n^{2}\right) / 2 \sigma^{2}}\end{equation}
where $\sigma$ is the standard deviation of the normal distribution. It refers to the variance that related to the Gaussian ﬁlter that reflects the degree to which the image is blurred. We regard DMI as the lowest layer of the Gaussian pyramid denoted as $G_1$. A set of images $\left\{ G_1, G_2, \dots,G_L \right\}$ generated by Eq. (2) constitute an $L$-layer Gaussian pyramid in which $G_{l+1}$ is $1/c^2$ size of $G_l$. Thus, a series of Gaussian pyramids represented as {$\rm GP_{L}$}-DMI are simply calculated by this iterative scheme. In this paper, we set $c$ to $2$ and utilize a $5\times5$ Gaussian kernel as Eq. (4). The pyramid algorithm reduces the filter band limit between layers by an octave, and chops the sampling interval by the same factor. The frequency of downsampling operations is related to the size of the original image. For the Gaussian pyramid based on an $M\times N$ image, the maximum number of layers is $\left\lfloor\log _{2}\min \{M, N\}\right\rfloor$.
\begin{equation}\varpi=\begin{bmatrix}
	1 & 4 & 6 & 4 & 1 \\
	4 & 16 & 24 & 16 & 4 \\
	6 & 24 & 36 & 24 & 6 \\
	4 & 16 & 24 & 16 & 4 \\
	1 & 4 & 6 & 4 & 1
\end{bmatrix}\end{equation}
\par Influenced by the complexity and concurrency of human behaviors, a simple action may also involve the movement of multiple body parts. We view the inherent characteristic inside the body as static information, while the contour information that can better describe the changing of movements as dynamic information. For the majority of actions, the static information inside the human body is highly similar. Take waving arms in different directions for instance, the information of the abdomen and legs are constant to some extent, and cannot provide the discriminative feature for recognition very well. On the contrary, the dynamic information of different body parts can better reflect the spatial changes of actions in the interval, thus reflecting the specific feature of this type of action. Inspired by this, we motivate to obtain the dynamic information of motions. We interpolate the $l_{th}$ layer of the Gaussian pyramid, that is, insert 0 in even rows and columns. Then, we utilize Gaussian filter to get $G_l^*$ which has the equal size as the image one layer below it. We caculate the difference between $G_l$ and $G_l^*$ to get the multi-scale dynamic infomation. At the same time, this operation removes a lot of static information, making LP-DMI more compact than GP-DMI. As in the Gaussian pyramid, we set c to 2. Mathematically:
\begin{gather}
G_{l}^{*}(i, j)=4 \sum_{m=-2}^{2} \sum_{n=-2}^{2} \varpi(m, n) \otimes G_{l}\left(\frac{i+m}{2}, \frac{j+n}{2}\right), \notag \\
\left(1 \leq l \leq L, 0 \leq i \leq R_{l}, 0 \leq j \leq C_{l}\right)
\end{gather}
\text{and}
\begin{equation}
G_{l}\left(\frac{i+m}{2}, \frac{j+n}{2}\right)=\left\{\begin{array}{ll}
G_{l}\left(\frac{i+m}{2}, \frac{j+n}{2}\right), & \text{if $\frac{i+m}{2}, \frac{j+n}{2}\in \mathbb{N^+}$ }\\
$ 0 $, & \text{otherwise }
\end{array}\right.
\end{equation}
\par Therefore, the Laplacian pyramid can be calculated as follows.\\
\begin{equation}
\left\{\begin{array}{ll}
L P_{l}=G_{l}-G_{l+1}^{*}, & 1 \leq l<L \\
L P_{L}=G_{L}, & l=L
\end{array}\right.
\end{equation}
where $LP_l$ is the $l_{th}$ layer of the Laplacian pyramid. Considering the integrity of motion infomation, we directly take the top layer of Gaussian pyramids as that of the Laplacian pyramid. Consequently, they have equal number of layers. Specifically, each depth frame produces three depth action images according to three views, thereby, it has three generated Laplacian pyramids. As shown in Fig. 4, the Laplacian pyramids cut down a large amount of static information inside the body meanwhile strengthen the dynamic information of body boundaries. This is more conducive to extracting discriminative features. In Sec. 4, we will further evaluate the proposed LP-DMI.
\begin{figure}[!t]
	\centering{\includegraphics[width=0.75\textwidth]{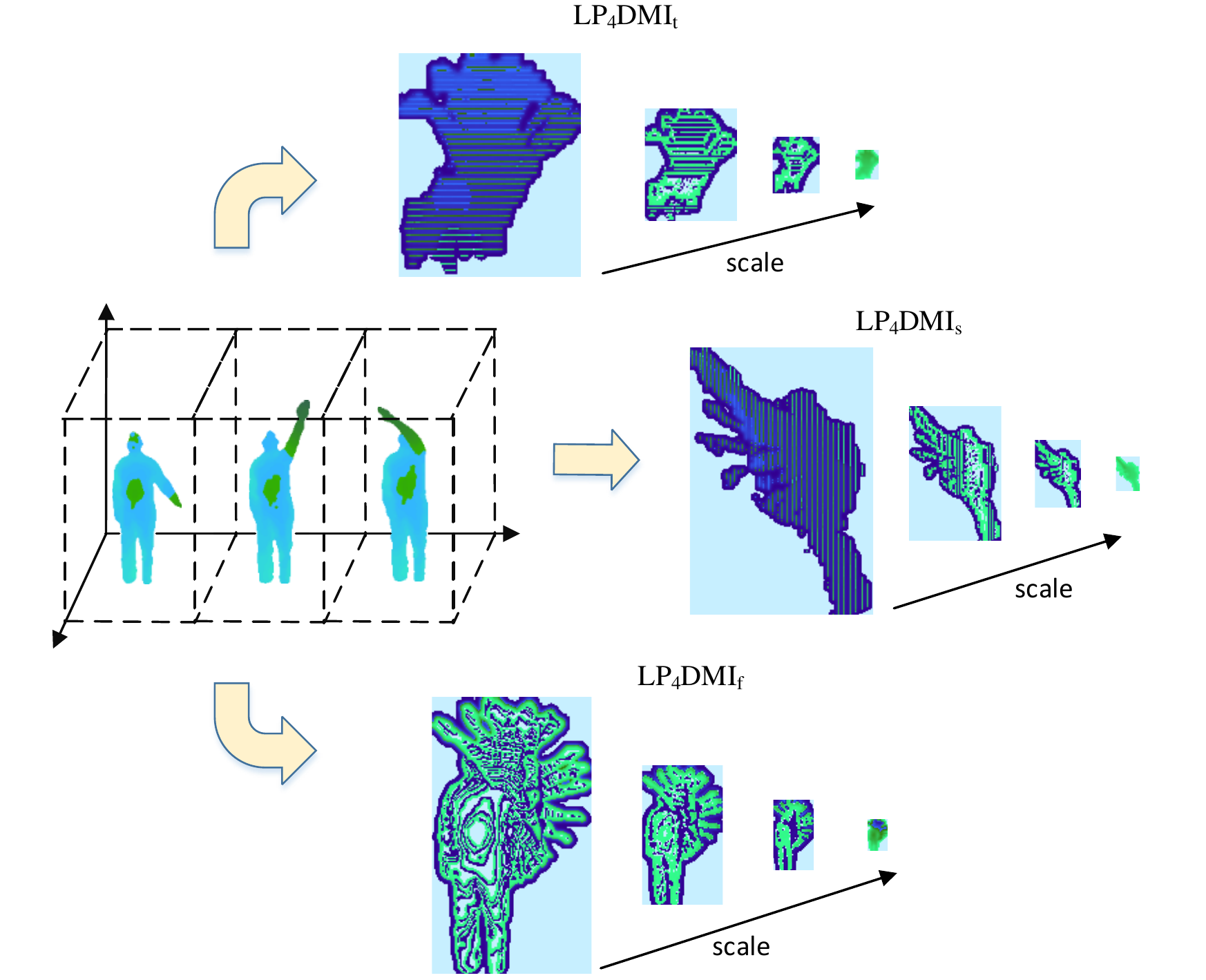}}
	\caption{An example of a four-layer $\rm {LP_4}$-DMI with three angles\label{fig4}}
\end{figure}
\subsection{Feature extraction and action classification}
\par There are several reasonable options for determining which feature to extract{\cite{RF54, RF55, RF56}}. In this paper, we utilize HOG descriptors to extract the local features of LP-DMI denoted as LP-DMI-HOG. HOG is sensitive to the distribution of gradient and edge information, thus it characterizes gradient changes especially the shape of objects pretty well. The basic idea is to compute gradient orientation histograms on a dense grid of uniformly spaced cells and perform local contrast normalization \cite{RF19}. Before extracting features, we copy adjacent pixels to normalizing the feature maps from the same view to the same size. The interpolated pixel values are the same as the neighboring pixels, so they will not interfere with the multi-scale information and we can compute multi-granularity motion features effectively. Moreover, this step is beneficial to solve the problem of too small pictures caused by incremental layers. We cascade the HOG feature extracted from LP-DMI in the same layer to obtain the three-view features at the same scale. Then we derive LP-DMI-HOG from coarse-grained to fine-grained as the layer increases. We normalize the resulting feature vectors using min-max scaling, and the principal component analysis (PCA) was applied to reduce the dimension for the sake of computational efficiency. Then, the resulting feature will be put into ELM for action classification.
\par The ELM algorithm was proposed by Huang $et$ $al$. for training single-hidden layer feed-forward neural networks (SLFNs) \cite{RF34}. The weight between the input layer and hidden layer can be initialized randomly as well as the bias of the hidden nodes. Therefore, the ELM just calculates the weight matrix between the hidden layer and output layer without the need to tune parameters. The matrix can be figured out by finding the generalized inverse matrix, thus the extreme learning machine has distinct advantages in parameter selection and computational efficiency. That is why we use extreme learning machine for action recognition. Given a training set with $n$ samples and $m$ classes $ D=\left\{\left(x_{i}, y_{i}\right) | x_{i} \in R^{n}, y_{i} \in \\ R^{m}, i=1,2, \ldots, n\right\} $, the SLFNs with $N$ hidden nodes can be expressed as:
\begin{equation}f\left(x_{i}\right)=\sum_{j=1}^{N} \beta_{j} g\left(w_{j} \cdot x_{i}+b_{j}\right)=o_{i}, i=1,2, \ldots, N\end{equation}
where $w_j=(w_{j1},w_{j2},...,w_{jd})^T$ is the weight vector connecting the $j_{th}$ hidden node with the input nodes. $\beta_{j}=(\beta_{j1},\beta_{j2},...,\beta_{jm})^T$ is the weight vector connecting the $j_{th}$ hidden node with the output nodes. $b_j$ represents the threshold of the $j_{th}$ hidden neuron, and $g(x)$ denotes the activation function. $w_j$ and $b_j$ are assigned randomly. The goal of ELM is to minimize the training error as far as possible, which can be depicted as $\sum_{i=1}^{N}\left\|o_{i}-y_{i}\right\|=0$. Therefore, parameters $\beta_j=(\beta_{j1},\beta_{j2},...,\beta_{jm})^T$ can be estimated by least-square ﬁtting with the given training data $D$. In other words, the problem can be written as the following equation.
\begin{equation} Y = H{\beta}\end{equation}
with\\
\begin{equation}H=\left(\begin{array}{ccc}
g\left(w_{1} \cdot x_{1}+b_{1}\right) & \dots & g\left(w_{m} \cdot x_{1}+b_{m}\right) \\
\vdots & \ddots & \vdots \\
g\left(w_{1} \cdot x_{n}+b_{1}\right) & \cdots & g\left(w_{m} \cdot x_{n}+b_{m}\right)
\end{array}\right)\end{equation}
$$\beta=\left(\beta_{1}^{T}, \beta_{2}^{T}, \ldots, \beta_{m}^{T}\right)^{T},$$
$$Y=\left(y_{1}^{T}, y_{2}^{T}, \ldots, y_{n}^{T}\right)^{T}$$
$H$ is the hidden layer output matrix of the network, in which the $j_{th}$ column is the $j_{th}$ hidden nodes output vector concerning inputs $ \left(x_1, x_2,\dots,x_m\right) $. The $i_{th}$ row of $H$ is the output vector of the hidden layer about input $x_i$. Once the input weight $w_j$ and the hidden layer bias $b_j$ are determined, the output matrix $H$ of the hidden layer is unique. The number of hidden nodes is usually much smaller than that of training samples. In this case, the smallest norm least-squares solution of Eq. (9) is equivalent to solving the following equation.
\begin{equation}\hat{\beta}=H^{\dagger} Y\end{equation}
where $H^{\dagger}$ is the Moore-Penrose generalized inverse of matrix $H$\cite{RF35}.
\section{EXPERIMENTS}
\par In order to evaluate the effectiveness of the proposed framework, we conduct experiments on the public MSRAction3D dataset{\cite{RF12}, UTD-MHAD{\cite{RF61}, and DHA dataset{\cite{RF11}. In Fig. 5, the depth video sequence of pickup and throw is shown as an example of action samples. We investigate how many layers are sufficient to capture multi-scale features for action recognition and compare several strategies of extracting local features. This section includes dataset description, experimental settings, parameter selection and results. 
\begin{figure}[t]
	\centering{\includegraphics[width=0.75\textwidth]{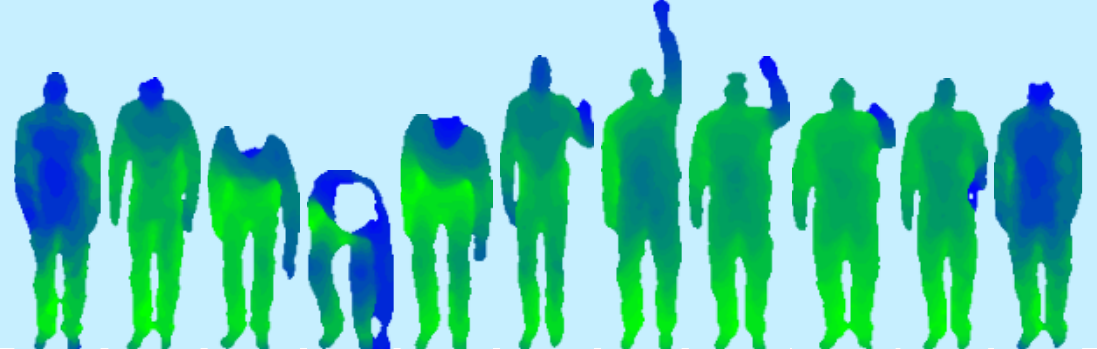}}
	\caption{The depth video sequences of pickup and throw in MSRAction3D dataset.\label{fig5}}
\end{figure}
\subsection{Datasets and experimental settings}
\par The MSRAction3D is a dataset for action recognition which contains 557 depth video sequences and 557 skeleton sequences for 20 actions captured by Kinect sensor. The actions including high arm wave, horizontal arm wave, hammer, hand catch, forward punch, high throw, draw x, draw tick, draw circle, hand clap, two hand wave, side boxing, bend, forward kick, side kick, jogging, tennis swing, tennis serve, golf swing, and pickup and throw are taken by 10 subjects. Every action is repeated by all the subjects two or three times. 
\par The UTD-MHAD is consists of 861 samples of 8 subjects. There are 27 actions in total, and every subject performed each action 4 times. The actions are: right arm swipe to the left, right arm swipe to the right, right hand wave, two hand front clap, right arm throw, cross arms in the chest, basketball shoot, right hand draw x, right hand draw circle (clockwise), right hand draw circle (counter clockwise), draw triangle, bowling, front boxing, baseball swing from right, tennis right hand forehand swing, arm curl, tennis serve, two hand push, right hand knock on door, right hand catch an object, right hand pick up and throw, jogging in place, walking in place, sit to stand, stand to sit, forward lunge, and squat.
\par The DHA database is orgnized with 483 depth video sequences for 23 actions. Each sample video was performed by 2 or 3 times by 21 subjects. The list of action classes are: bend, jack, jump, pjump, run, side, skip, walk, one-hand-wave, two-hand-wave, front-clap, side-clap, arm-swing, arm-curl, leg-kick, leg-curl, rod-swing, golf-swing, front-box, side-box, tai-chi, pitch, and kick. 
\par We conduct experiments with the following experimental settings. Setup1: In order to have fair experimental results, we perform the cross-subject tests on the three benchmark datasets according to the experimental settings of \cite{RF12,RF53}. More precisely, we use odd subjects for training, whereas even subjects are applied for testing.
\begin{table}[h]
	\caption{Three subsets of the MSRAction3D dataset}
	\label{table_1}
	\centering
	\begin{tabular}{cccccc}
		\hline\noalign{\smallskip}
		\bfseries Label&AS1&Label&AS2&Label&AS3\\
		\noalign{\smallskip}\hline\noalign{\smallskip}
		2&Horizontal arm wave&1&High arm wave&6&High throw\\
		3&Hammer&4&Hand catch&4&Forward kick\\
		5&Forward punch&7&Draw x&15&Side kick\\
		6&High throw&8&Draw tick&16&Jogging\\
		10&Hand clap&9&Draw circle&17&Tennis swing\\
		13&Bend	&11&Two hand wave&18&Tennis serve\\
		18&Tennis serve&14&Forward kick&19&Golf swing\\
		20&Pickup and throw&12&Side boxing&20&Pickup and throw\\
		\noalign{\smallskip}\hline
	\end{tabular}
\end{table} 
\par Setup2: We divide the MSRAction3D dataset into three subsets as shown in Table 1. Three different tests are conducted on these subsets following the settings as \cite{RF11}. In test 1, 1/3 action samples in each subset are employed as the training set, and the remaining samples are used for validation. On the contrary, test 2 uses 2/3 samples for training, and the rest samples are taken in the testing set. Test 3 has a cross-subject test on each subset abide by setup 1, that is to say, the action samples corresponding to the odd subjects in each subset are used for training and the rest for testing.
\subsection{Experimental results and discussion}
\begin{figure}[!t]
	\centering{\includegraphics[width=1.0\textwidth]{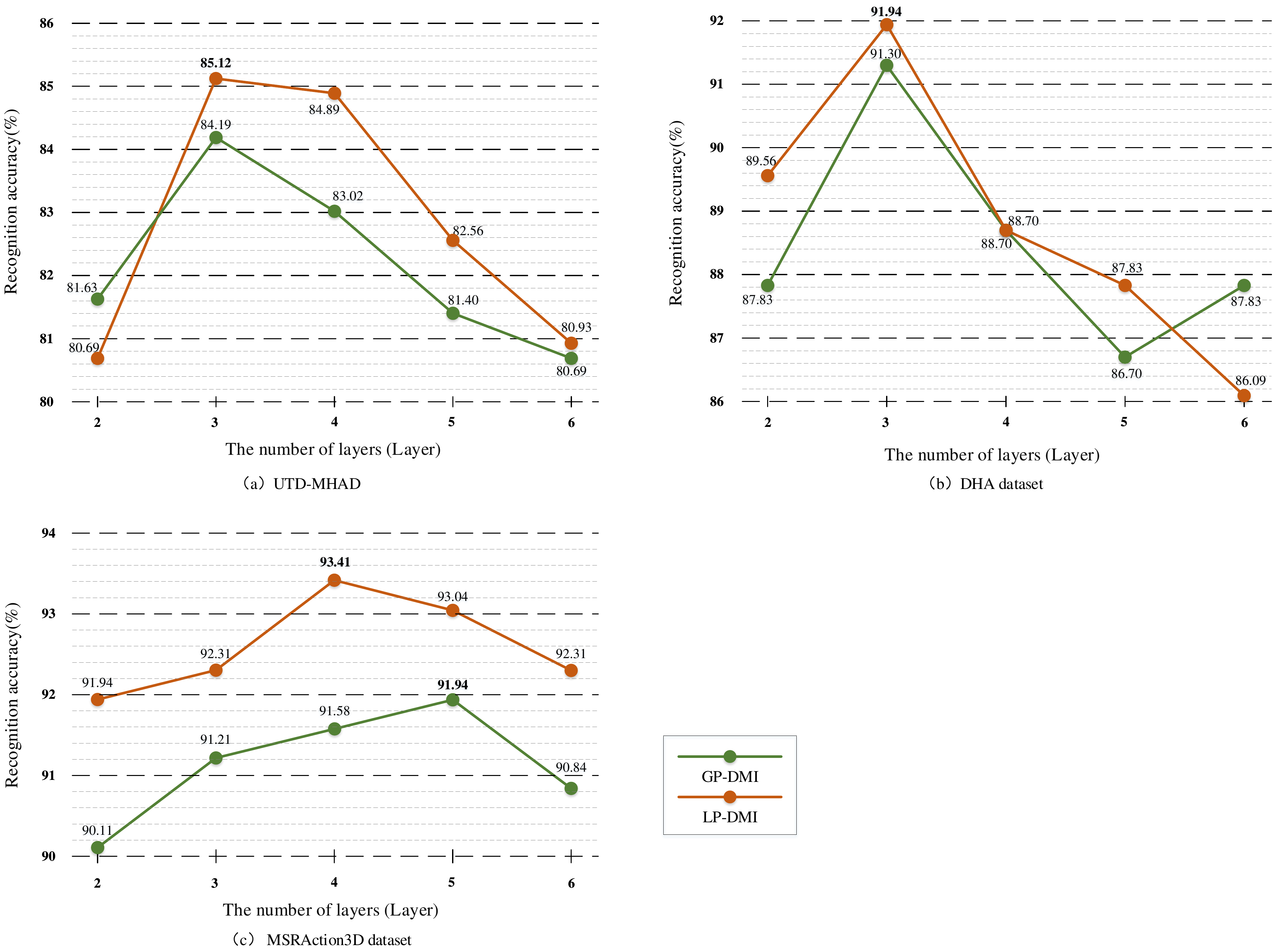}} 
	\caption{The recognition accuracy of LP-DMI with different layers\label{fig6}}
\end{figure} 
\begin{table}[t]  
	\renewcommand{\arraystretch}{1.5}
	\caption{The results of various normalization strategies on MSRAction3D dataset }
	\label{table_2}  
	\centering
	\setlength{\tabcolsep}{1.7mm}
	\begin{tabular}{c|p{1.5cm}p{1.5cm}p{1.5cm}|p{1.5cm}p{1.5cm}p{1.5cm}}  
		\hline
		\multirow{2}*{Datasets}&\multicolumn{3}{c|}{$ N_1 $}&\multicolumn{3}{c}{$ N_2 $} \\ \cline{2-7}
		& Accuracy(\%) & Dimension &Time(s) &Accuracy(\%) & Dimension &Time(s)  \\ 
		\hline
		{$ C_1 $}&89.01& 15444 & 91.25&$\mathbf{91.58}$&49968&254.88\\
		{$ C_2 $}&88.64& 49968&205.34&87.55&1234512&3151.40 \\
		\hline
	\end{tabular}
\end{table}

\begin{table}[t]  
	\renewcommand{\arraystretch}{1.5}
	\caption{The results of various feature extraction strategies on UTD-MHAD }
	\label{table_3}  
	\centering
	\setlength{\tabcolsep}{1.7mm}
	\begin{tabular}{c|p{1.5cm}p{1.5cm}p{1.5cm}|p{1.5cm}p{1.5cm}p{1.5cm}}  
		\hline
		\multirow{2}*{Datasets}&\multicolumn{3}{c|}{$ N_1 $}&\multicolumn{3}{c}{$ N_2 $} \\ \cline{2-7}
		& Accuracy(\%) & Dimension &Time(s) &Accuracy(\%) & Dimension &Time(s)  \\ 
		\hline
		{$ C_1 $}&79.53& 15444 & 120.36&$\mathbf{84.91}$&37476&255.38\\
		{$ C_2 $}&77.91& 37476&205.34&87.55&296676&272.25 \\
		\hline
	\end{tabular}
\end{table}
\begin{table}[t] 
	\renewcommand{\arraystretch}{1.5}
	\caption{The comparsion of different descriptors}
	\label{table_4}
	\centering
	\setlength{\tabcolsep}{1.7mm}
	\begin{tabular}{c|p{1.5cm}p{1.5cm}|p{1.5cm}p{1.5cm}}
		\hline
		\multirow{2}*{Datasets}&\multicolumn{2}{c|}{VGG}&\multicolumn{2}{c}{HOG} \\ \cline{2-5}
		& Accuracy(\%) &Time(s)& Accuracy(\%) &Time(s) \\
		\hline
		MSR& 91.94 & 14902.57&$\mathbf{93.41}$&125.35\\
		UTD& 81.86 & 10865.87&$\mathbf{85.12}$&142.43 \\
		DHA& 84.35 & 9387.73&$\mathbf{91.94}$&57.55 \\
		\hline
	\end{tabular}
\end{table}
\subsubsection{Parameter selection}
\par We normalize the action feature maps projected onto the same planes to a uniform size as shown in Fig. 2. We set the size of each cell to $10 \times 10$ pixels and the number of gradient orientation bins is $9$. The size of block is $2 \times 2$. Furthermore, the step is $10$ pixels. The remained principal components of MSRAction3D, UTD-MHAD, and DHA is 550, 450 and 860. So that, each feature vector is a total of $15444$ and $20592$ dimensions when the number of layers is 3 and 4 respectively. Note that, we consider this as the default setting of feature extraction.
\par To exploit the optimal muti-scale feature map of different datasets, we construct LP-DMI with different layers in a step-wise manner and do experiments according to setup1. The experimental results with respect to the GP-DMI and LP-DMI from 2 to 6 layers are presented in Fig. 6. The first thing we noticed is that the motion feature will be coarse-grained and incomplete for recognition if the number of layers is inadequate. Otherwise, if the number of layers is superfluous, the static information may be more redundant, which leads to low efficiency and accuracy. In addition, the experimental results illustrate that LP-DMI yeilds better recognition accuracy on the whole, which achieves the highest recognition rate of 93.41\% when the number of layers is 4 on MSRAction3D dataset. The $\rm LP_3$-DMI on UTD-MHAD and DHA dataset are the optimum, and the recognition rates are respectively 85.12\% and 91.94\%. We will abide by the optimal layer setting obtained here in subsequent experiments.
\par After that, we compare several strategies of feature extraction and normalization. In order to prevent the size of feature maps and cells from being too small, we adopt a combination of two dynamic constraint in this experiment rather than the default experimental setting. ${D^l}_v(w, h, d)$ is the normalization parameter denoting that the size of $LP_l$-$ DMI_f$, $LP_l$-$ DMI_s$, $LP_l$-$ DMI_t$ is $ w \times h$, $ h \times d$, $ w \times d$. Constraint $N_1$: ${D^l}_v(w/2^{l-1}, h/2^{l-1}, d/2^{l-1})$. Constraint $N_2$: ${D^l}_v(w, h, d)$ = ${D^l}_v(160, 320, 240)$. Constraint $C_1$: the size of cell is $20 \times 20 $. Constraint $C_2$: the size of cell is $(20/2^{l-1}) \times (20/2^{l-1})$. The experimental results are reproted in Table 2 and Table 3. We observe that $N_2$ combined with $C_1$ outperforms other strateges. In other words, the applied normalization method achieve the effect of improving the classification accuracy. We further illustrate that the LP-DMI-HOG descriptor is more efficient and more discriminative than the descriptor extracted by VGG-16{\cite{RF56}} as depicted in Table 4.
\begin{table}[t]
	\caption{The comparation of other feature maps on MSRAction3D dataset(\%) }
	\label{table_5}
	\centering
	\setlength{\tabcolsep}{1.7mm}
	\begin{tabular}{ccccccccc}
		\hline\noalign{\smallskip}
		{\quad}&{\quad}&{MEI}&{MHI}&{DMM}&{HP-DMM}&{DMI}&{$\rm GP_5$-DMI}&{$\rm LP_4$-DMI}\\ 
		\noalign{\smallskip}\hline\noalign{\smallskip}
		\multirow{4}{*}{Test One}
		&AS1&75.34&67.81&92.47&89.73&89.04&93.84& $\mathbf{95.21}$\\
		&AS2&71.05&69.74&84.21&80.92& $\mathbf{86.84}$&86.18&86.18\\
		&AS3&71.62&70.95&86.49&84.46&85.81&86.49& $\mathbf{89.86}$\\
		&Average&72.67&69.50&87.72&85.04&87.23&88.84& $\mathbf{90.42}$\\
		\multirow{4}{*}{Test Two}
		&AS1&89.04&84.93& $\mathbf{97.26}$& $\mathbf{97.26}$&93.42&93.42& $\mathbf{97.26}$\\
		&AS2&82.89&82.89&86.84&88.16&93.42&93.42& $\mathbf{98.63}$\\
		&AS3&90.54&93.24&97.30&94.59&97.30&97.30& $\mathbf{100.00}$\\
		&Average&87.49&87.02&93.80&93.34&94.71&94.71& $\mathbf{98.63}$\\
		\multirow{4}{*}{Test Three}
		&AS1&71.43&72.38& $\mathbf{99.05}$&91.43&94.29&97.14&98.10\\
		&AS2&69.64&66.07&85.71&84.82&82.14&89.29& $\mathbf{90.18}$\\
		&AS3&74.78&70.27&94.59&92.79&93.69&92.79& $\mathbf{95.50}$\\
		&Average&71.95&69.57&93.12&89.68&90.04&93.07& $\mathbf{94.59}$\\
		\noalign{\smallskip}\hline
	\end{tabular}
\end{table}
\begin{table}[t]
	\caption{The recognition rate of depth feature maps on UTD-MHAD and DHA dataset(\%) }
	\label{table_6}
	\centering
	\begin{tabular}{cccccccc}
		\hline\noalign{\smallskip}
		{\quad}&{MEI}&{MHI}&{DMM}&{HP-DMM}&{DMI}&{$\rm GP_3$-DMI}&{$\rm LP_3$-DMI}\\ 
		\noalign{\smallskip}\hline\noalign{\smallskip}
		UTD&51.63&56.52&74.19&77.67&76.28&84.19& $\mathbf{85.12}$ \\
		DHA&62.60&79.10&90.4&91.30&90.80&91.30&$\mathbf{91.94}$ \\
		\noalign{\smallskip}\hline
	\end{tabular}
\end{table}
\begin{figure}[t]
	\centering{\includegraphics[width=1.0\textwidth]{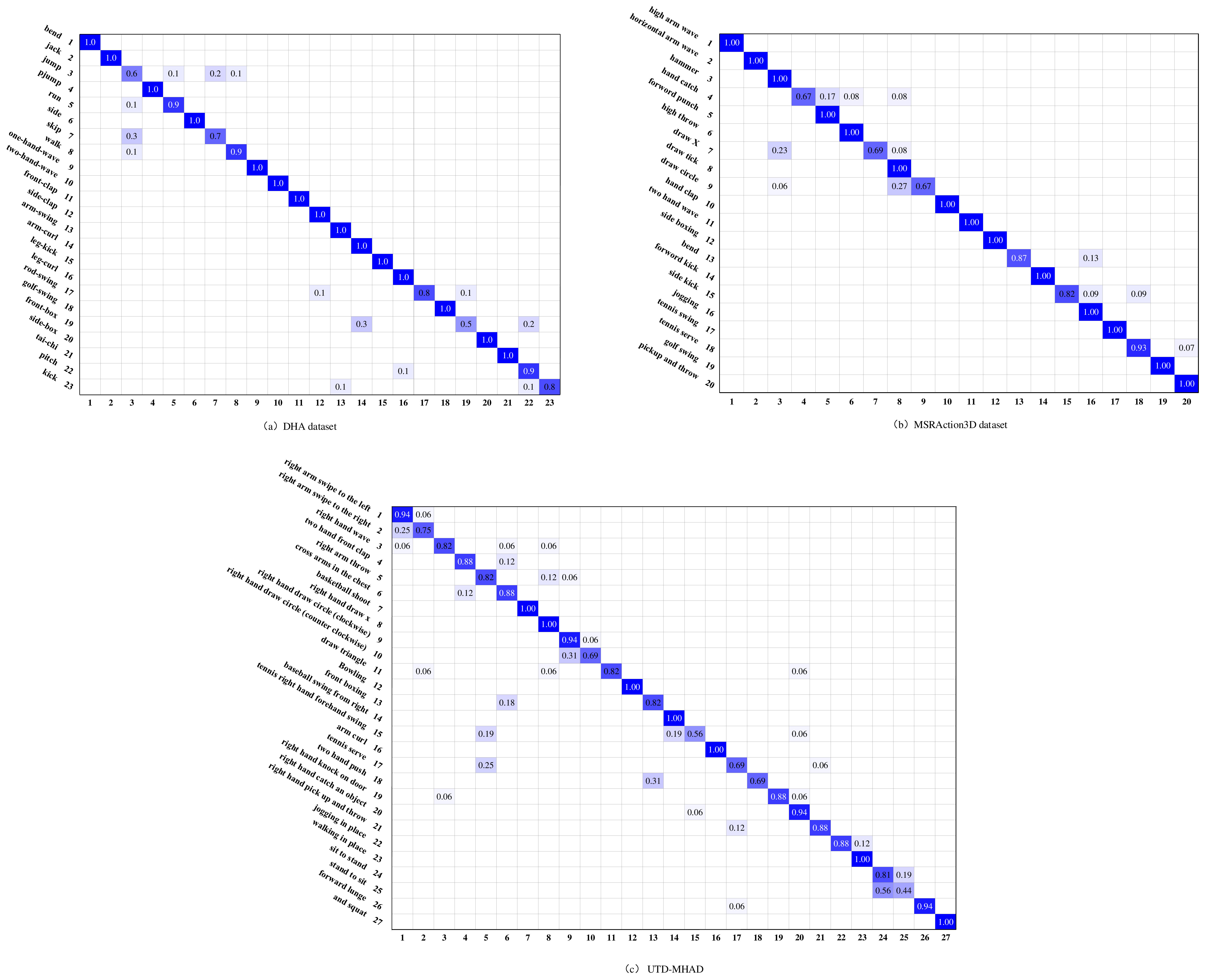}}
	\caption{The confusion matrix of three datasets \label{fig7}}
\end{figure}
\subsubsection{Evaluation of LP-DMI}
\par We evaluate the efficiency of LP-DMI from two aspects. On the one hand, we have proved that LP-DMI is a more discriminative multi-scale feature map compared with GP-DMI. On the other hand we will certify that the LP-DMI-HOG extracted from LP-DMI excels HOG features based on other feature maps. For the fairness of the results, we follow our parameter settings on these depth maps to obtain HOG descriptor. At the same time, we all employ ELM for action recognition. We conduct experiments on MSRAction3D dataset following setup2. The alone and average results with regard to AS1, AS2, and AS3 are presented in Table 5, and the highest accuracy of each subset has been shown in bold. As can be seen, LP-DMI achieves the highest average recognition rate in three different tests and outperforms than other feature maps. Specifically, in test one, LP-DMI achieved 90.42\% accuracy on the three subsets. In addition to the performance on AS2 which is slightly lower than DMM, LP-DMI has an absolute advantage on other two subsets. In the second test, our proposal exceeds others significantly and gets the best recognition rate of 98.63\% on AS2. Furthermore, the ELM trained by LP-DMI-HOG even can completely label all the testing samples on AS3. Therefore, in spite of the recognition rate of DMM and HP-DMM on AS1 equals to our method, the average recognition rate we have achieved is still 5\% higher than them. In test three, LP-DMI obtains an average recognition rate of 94.59\%. The result of LP-DMI on AS1 is 0.95\% mildly lower than that of the DMM, but the recognition rates on other subsets are optimal. Overall, LP-DMI surpasses MEI, MHI and {$\rm GP_{5}$}-DMI in all tests. Although DMM, HP-DMM, and DMI on individual subsets are superior to LP-DMI, the average recognition rate of our method is the highest. It should be noted that we almost improved accuracy by 4\% in three tests by constructing Laplacian Pyramid pyramid for DMI, and this transformation process is very efficient and does not cause too much time consumption. 
\par On UTD-MHAD and DHA dataset, we testify the proposed complying with setup1, and describe the result in Table 6. Our method is shown in bold. It can be seen that $\rm LP_3$-DMI yeilds the best recognition accuracy of 85.12\% on UTD-MHAD. Once more, the experiments of DHA dataset validate our methods in which $\rm LP_3$-DMI produce the result of 91.94\%. It is worth noting that our method significantly improves the recognition rate compared with the original feature map DMI. For elaborating the performance of our method clearly, the confusion matrix computed from three datasets is depicted in Fig. 7. It can be seen that our method can correctly recognize the majority of actions. After analyzing the accuracy of specific classes, we find that the errors mainly occur in the classification of similar actions. For example, skip and jump, front-box and arm-curl, draw x and draw tick. In a word, this experimrnt further confirms that LP-DMI is a compact multi-scale feature map, and the proposed LP-DMI-HOG descriptor is promising.
\begin{table}[t]
	\caption{Comparison of our method with baseline methods on MSRAction3D dataset}
	\label{table_7}
	\centering
	\begin{tabular}{ccc}
		\hline\noalign{\smallskip}
		Methods&Modality&Accuracy(\%)\\
		\noalign{\smallskip}\hline\noalign{\smallskip}
		Pose Set\cite{RF38}&D&90.0\\  
		DMM-HOG\cite{RF19}&D&88.7\\
		DMM-GLAC\cite{RF28}&D&88.9\\
		HON4D\cite{RF25}&D&88.9\\
		Skeletons Lie group\cite{RF40}&S&89.5\\ 
		HOG3D+LLC\cite{RF41}&D&90.9\\
		3D-CNN\cite{RF42}&D&86.1\\
		DSTIP\cite{RF15}&D&89.3\\
		STOP \cite{RF44}&D&87.5\\
		HP-DMM-CNN\cite{RF21}&D&92.3\\
		PointLSTM-late\cite{RF64}&D+S+RGB&95.3\\  
		MMHCCCA\cite{RF21}&D+S&93.5\\
		Ji $et$ $al$.\cite{RF43}&D+S&90.8\\
		PointLSTM-late\cite{RF64}&D+S&95.3\\
		Trelinski $et$ $al$.\cite{RF48}&D&90.6\\
		Ahmad $et$ $al$.\cite{RF46}&D&87.1\\
		Wei $et$ $al$.\cite{RF49}&S&87.2\\
		Xin $et$ $al$.\cite{RF65}&D+S&91.5\\ 
		LP-DMI&D&$\mathbf{93.4}$\\
		LP-DMI+HP-DMM&D&$\mathbf{94.8}$\\
		LP-DMI+MJD&D+S&$\mathbf{95.6}$\\
		LP-DMI+ST-GCN&D+S&$\mathbf{94.5}$\\
		\noalign{\smallskip}\hline
	\end{tabular}
\end{table}
\subsubsection{Comparison with baseline methods}
\par In this experiment, we follow setup1 as the baseline methods for persuasion. The cross-subject test is challenging due to variations in the same actions performed by different subjects, but our method can still achieve high accuracy. In Table 7, our method obtains the highest accuracy of 93.4\% compared with other solutions utilizing single depth modality data on MSRAction3D dataset, and it is 4.5 percentage points higher than DMM-GLAC which extracts local feature descriptor from depth motion maps as well. The HP-DMM-CNN, 3D-CNN as well as the method in \cite{RF46} using convolutional neural networks are 1.1, 7.3, 9.9 points lower than our method. It should be noted that the method proposed by Ji $et$ $al.$ \cite{RF43} is 2.6 points lower than ours although they obtain local spatio-temporal scaled pyramid and embed skeleton information. Furthermore, we fuse LP-DMI-HOG descriptor with HP-DMM-HOG, MJD-HOG ,ST-GCN\cite{RF57}, which yeilds the recognition rate of 94.8\%, 95.6\%, 94.5\%. We also demonstrate the generality of our framework on UTD-MHAD and report the results in Table 8. With same evaluation strategy, we compare our system with depth-based and multi-modal feature fusion methods. Notably, our method is advantageous with the accuracy of 85.1\% which is 2.4, 3.6, 11.4 points higher than HP-DMM-CNN, DMM-HOG and HP-DMM-HOG. Overall, our approach is encouraging and is able to have superior performance by the means of fusion technologies.
\begin{table}[t]
	\caption{Comparison of our method with baseline methods on UTD-MHAD}
	\label{table_8}
	\centering
	\begin{tabular}{ccc}
		\hline\noalign{\smallskip}
		Methods&Modalities&Accuracy(\%)\\
		\noalign{\smallskip}\hline\noalign{\smallskip}
		DMM-HOG\cite{RF19}&D&81.5\\
		3DHOT-MBC\cite{RF62}&D&84.4\\
		Hierarchical Gaussian\cite{RF66}&D&84.1\\
		HP-DMM-CNN\cite{RF21}&D&82.7\\
		HP-DMM-HOG\cite{RF67}&D&73.7\\
		MLSL\cite{RF60}&D+S&88.3\\
		Kamel $et$ $al$.\cite{RF12}&D+S&88.1\\
		Chen $et$ $al$.\cite{RF61}&D+S+RGB&79.1\\
		STSDDI.\cite{RF63}&D+S+RGB&91.1\\
		LP-DMI&D&$\mathbf{85.1}$\\
		LP-DMI+HP-DMM&D&$\mathbf{89.0}$\\
		LP-DMI+MJD&D+S&$\mathbf{90.4}$\\
		LP-DMI+ST-GCN&D+S&$\mathbf{94.1}$\\
		\noalign{\smallskip}\hline
	\end{tabular}
\end{table}
\section{Conclusion}
\par In this paper, we proposed a novel method based on the Laplacian pyramid considering multi-scale information for human action recognition. We calculated LP-DMI to increase the scale diversity of depth motion images in order to capture the multi-scale motion features and strengthen more favorable dynamic information. The experiments conducted on MSRAction3D, UTD-MHAD and DHA dataset have demonstrated that our approach outperforms baseline methods. The LP-DMI is more compact and discriminative than existing feature maps. Furthermore, the extracted LP-DMI-HOG which contains multi-granularity features has effectively improved the accuracy of action recognition. However, our method is still flawed in identifying actions with similar motion trajectories. The future work will focus on fusing multimodal features and considering multi-scale temporal information to facilitate the recognition accuracy. 

\end{document}